\documentclass[12pt]{opt2019} 

\usepackage{subcaption}
\usepackage{floatrow}

\title[Using Dimensionality Reduction to Optimize t-SNE]{Using Dimensionality Reduction to Optimize t-SNE}

\optauthor{\Name{Rikhav Shah} \Email{rikhav.shah@berkeley.edu}\\
	\addr UC Berkeley, Berkeley, CA 94720
	\AND
  \Name{Sandeep Silwal} \Email{silwal@mit.edu}\\
  \addr Massachusetts Institute of Technology, Cambridge, MA 02139}

\begin{document}

\maketitle

\begin{abstract}%
t-SNE is a popular tool for embedding multi-dimensional datasets into two or three dimensions.  However, it has a large computational cost, especially when the input data has many dimensions.  Many use t-SNE to embed the output of a neural network, which is generally of much lower dimension than the original data.  This limits the use of t-SNE in unsupervised scenarios.  We propose using \textit{random} projections to embed high dimensional datasets into relatively few dimensions, and then using t-SNE to obtain a two dimensional embedding.  We show that random projections preserve the desirable clustering achieved by t-SNE, while dramatically reducing the runtime of finding the embedding.
\end{abstract}


\section{Introduction}\label{sec:intro}
Cluster analysis of high dimensional datasets is a core component of unsupervised learning and a common task in many academic settings and practical applications. A widely used practice to more easily visualize the cluster structure present in the data is to first reduce the dimension of data to a lower dimension, typically two. A popular tool to perform this dimensionality reduction is the t-distributed Stochastic Neighborhood Embedding (t-SNE) algorithm, introduced by van der Maaten and Hinton \cite{orig_tsne}. t-SNE has been empirically shown to preserve local neighborhood structure of points in many applications and the popularity of this method can be seen by the number of citations of the original paper ($9000+$) and implementations in popular packages such as scikit-learn.

The main drawback of t-SNE however, is its large computational cost, a big part of which can be written as $O($\# of distances computed $\cdot d)$ where $d$ is the dimension of the input points.  This drawback severely limits its potential use in unsupervised learning problems. To combat the large computational cost of t-SNE, many practical developments have been made using techniques such as parallel computation, GPU, and multicore implementations with a large chunk of work on reducing the number of pairwise distances that are computed between the input points \cite{gpu1, gpu2, treebased}. We take a different approach by focusing instead on reducing the value of $d$ in the runtime cost.

In this work, we show that the distance computation can be optimized by first projecting the input data into a much smaller random subspace before running t-SNE. Our motivation for this approach comes from the field of metric embeddings where it is known that random projections preserve many geometric structures of the input data. For more information, see Section \ref{sec:motivation}. More formally, our contributions are the following:
\begin{itemize}
    \item We empirically show that projecting to a very small dimension (even as low as $5\%$ of the original dimension in some cases) and then performing t-SNE preserves local cluster information just as well as t-SNE with no prior dimensionality reduction step.
    \item We empirically show that performing dimensionality reduction first and then running t-SNE is significantly faster than just running t-SNE.
\end{itemize}
We note that our last result is implementation agnostic, since every implementation must compute distances among some of the input points. To show the benefits of our approach and simplify our experiments, we use two implementations of t-SNE: a widely available, but slower scikit-Learn implementation, and a highly optimized Python implementation called `openTSNE' \cite{opentsne}. For more details about our experimental setup, see Section \ref{sec:setup}.

\subsection{Measuring Quality of an Embedding}
We devise an \emph{accuracy score} to measure the quality of the cluster structure of an embedding. First we note that ideal clusters are ones in which every element has the same label as others in the same cluster. If one has labels for only a few datapoints, this kind of clustering would allow much of the dataset to be correctly labeled. Thus ideally, such clustering would be created without using labels (note that neither random projections nor t-SNE use labels to determine embeddings). A datapoint is said to be ``correctly clustered" in the low-dimension embedding if it's label matches that of its nearest neighbor. The \emph{accuracy score} of an embedding is given by the fraction of datapoints that are correctly clustered. Such a score could be refined by instead considering the modal label of the $k$-nearest-neighbors. The accuracy score rewards the case that clusters in high dimension stay together in the lower dimension and penalizes the case of different clusters in the higher dimension merging together in the lower dimension since then we can expect the nearest neighbors of many points to have a differing label. In Section \ref{sec:actualresults}, we compare the accuracy scores of performing dimensionality reduction and then t-SNE versus performing only t-SNE on various labeled datasets.

\subsection{Overview of t-SNE}\label{sec:overview}
Given a set of $N$ points $x_1, \cdots, x_N \in \mathbb{R}^d$, t-SNE first computes the similarity score $p_{ij}$ between $x_i$ and $x_j$ defined as $p_{ij} = (p_{i \mid j} + p_{j \mid i})/(2N)$ where for a fixed $i$, $p_{j \mid i} \propto \exp(\|x_i-x_j\|^2/\sigma_i^2)$ for some parameter $\sigma_i$. Intuitively, the value $p_{ij}$ measures the `similarity' between points $x_i$ and $x_j$. t-SNE then aims to learn the lower dimensional points $y_1, \cdots, y_N \in \mathbb{R}^2$ such that if $q_{ij} \propto (1+\|y_i-y_j \|_2^2)^{-1}$, then $q_{ij}$ minimizes the Kullback–Leibler divergence of the distribution $\{q_{ij}\}$ from the distribution $\{p_{ij}\}$. For a more detailed explanation of the t-SNE algorithm, see \cite{orig_tsne}.

\subsection{Motivation for Using Random Projections}\label{sec:motivation}
Like t-SNE, using a random projection is also a dimensionality reduction tool which has been shown to preserve many geometrical properties of the input data, such as pairwise distances, nearest neighbors, and cluster structures \cite{jlproof, nearestneighbor_embeddings, kclustering}. One of the key results of this field is the well known Johnson–Lindenstrauss (JL) Lemma which roughly states that given any set of $N$ points, a random projection of these points into dimension $O(\log N/\epsilon^2)$ preserves all pairwise distances upto multiplicative error $(1 \pm \epsilon)$ \cite{jlproof}. However, it is not possible to use a random projection to project down to a very small subspace, such as two dimensions used in most applications, while still maintaining the desired geometric properties. This is in contrast to t-SNE which has been empirically shown to preserve many local properties of the input data even in very low dimensions (such as two). However, we can leverage advantages of both of these approaches through the counter intuitive idea of using dimensionality reduction \emph{before} dimensionality reduction. More specifically, we can hope that by first randomly projecting the input data into a smaller dimension, we can preserve enough local information for t-SNE to use to be able to meaningfully project the data into two dimensions and still maintain the inherent cluster structure. 

Indeed, if we consider the similarity scores $p_{ij}$ computed by t-SNE, we see that if a random projection preserves pairwise distances, then the values of the similarity scores remain the same. Furthermore, if we require the weaker condition that the random projection preserves nearest neighbor relations, then we see that the values $p_{ij }$ maintain their order. More precisely, if $\|x_i - x_j\| < \|x_i - x_k\|$ holds, the similarity scores between points $x_i$ and $x_j$ will be large and thus, t-SNE will try to place these points closer together. Using a random projection from $\mathbb{R}^d$ to $\mathbb{R}^k$, we immediately get a reduction of the runtime from $O($\# of distances computed $\cdot d)$ to $O($\# of distances computed $\cdot k)$.  We show the success of our method in Section \ref{sec:results}.

\subsection{Related Works}
There have been many works about improving t-SNE through better implementations, such as GPU acceleration or parallel implementations \cite{gpu1, gpu2, multicoretsne}. There has also been work on reducing the number of distances that need to be computed using tree based methods, see \cite{treebased} for more info. To our knowledge, our work is the first to suggest using random projections in conjunction with t-SNE. We have verified this claim by going through the papers that cite the original t-SNE paper of \cite{orig_tsne} and searching Google Scholar for papers with the phrase `random projection'.

\section{Experimental Results}\label{sec:results}
\subsection{Experimental Setup and Justification}\label{sec:setup}
We construct embeddings by first applying a random projection into $d'$ dimensions, then using t-SNE to arrive at a two dimensional embedding.  The random projections are performed by using an appropriately sized random matrix with i.i.d.\@ Gaussian entries $\mathcal{N}(0,1/d)$ where $d$ is the original dimension of the input data.  We repeat this process for several values of $d'$ exponentially spaced and ranging from $7$ to $d$.

We use the following four datasets in our experiments: MNIST, Kuzushiji-MNIST (KMnist) (data set of Japanese cursive characters in the same format as MNIST) \cite{kmnist}, Fashion-MNIST (data set of fashion images in the same format as MNIST) \cite{fashionmnist}, and the Small Norb data set (data set of images of toys) \cite{smallnorb}. All of our experiments were done in Python $3.7$ using one core of a MacBook Pro using a $2.7$ GHz Intel Core i$5$ processor. We only report the time to perform t-SNE since the t-SNE runtime is many orders of magnitude larger than all the other steps. Our code can be accessed from the following GitHub reprository: \url{https://github.com/ssilwa/optml}. 

\subsection{Results}\label{sec:actualresults}
\begin{figure}[htb]
\begin{center}
\includegraphics[width=.4\textwidth]{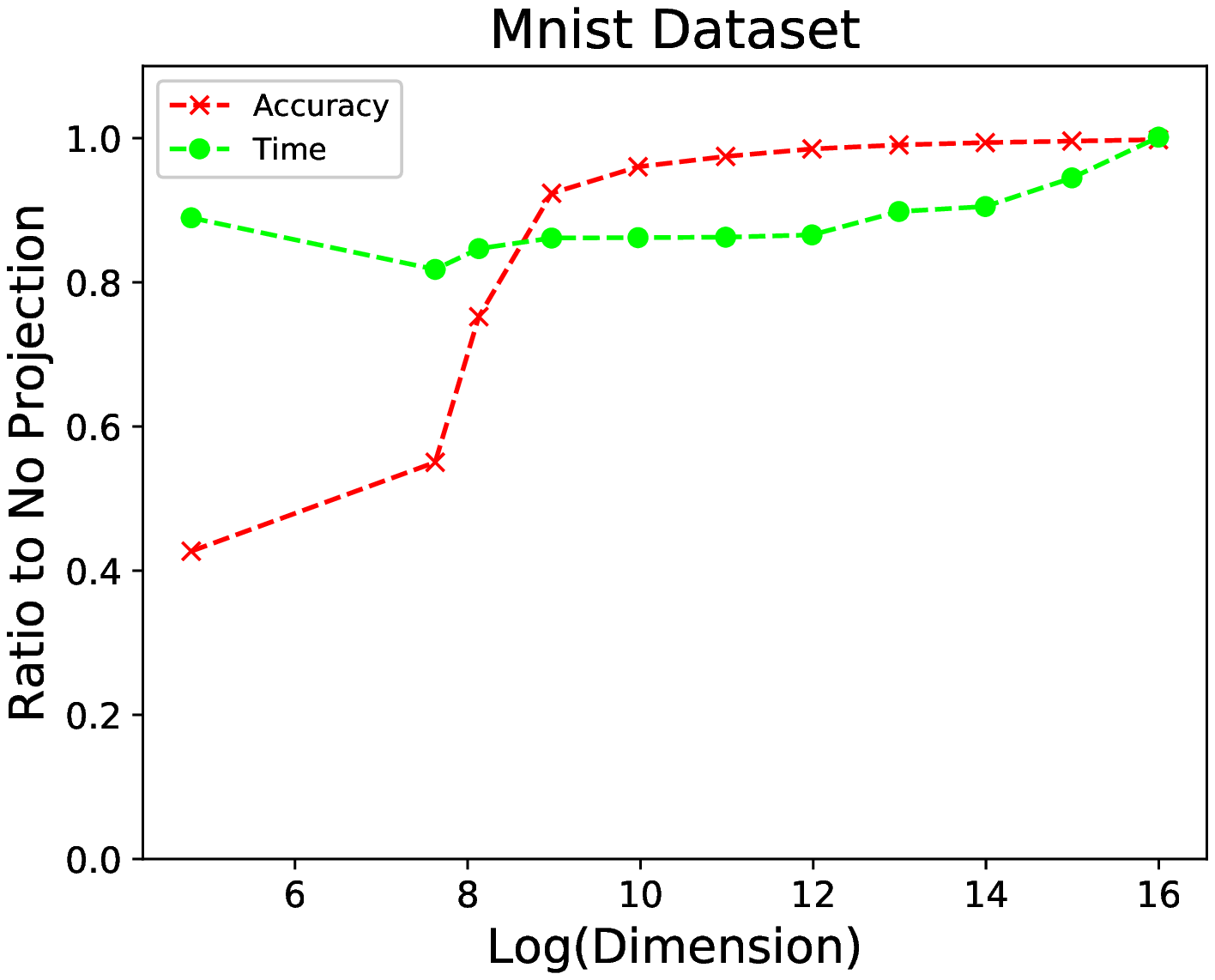}
\includegraphics[width=.4\textwidth]{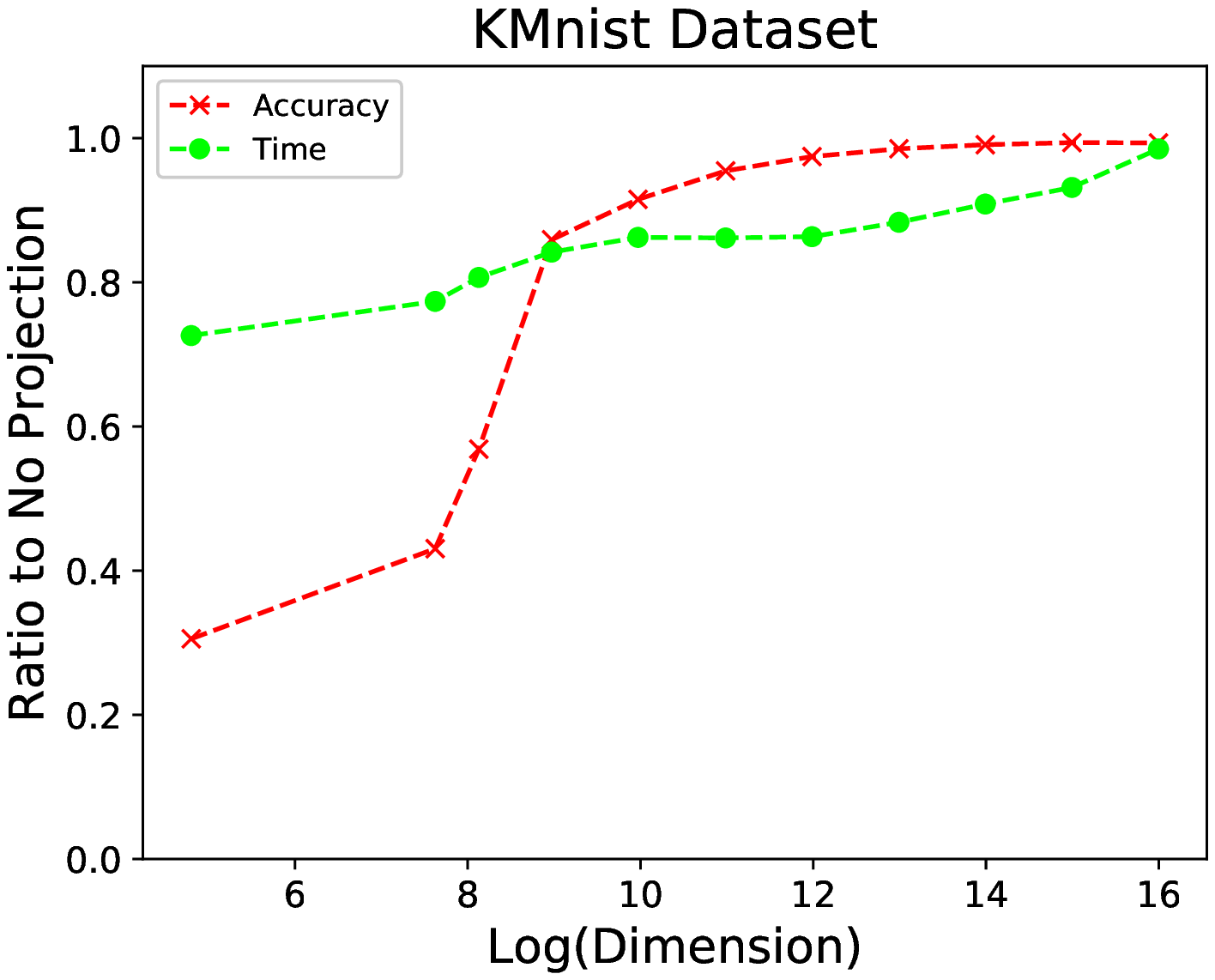}
\includegraphics[width=.4\textwidth]{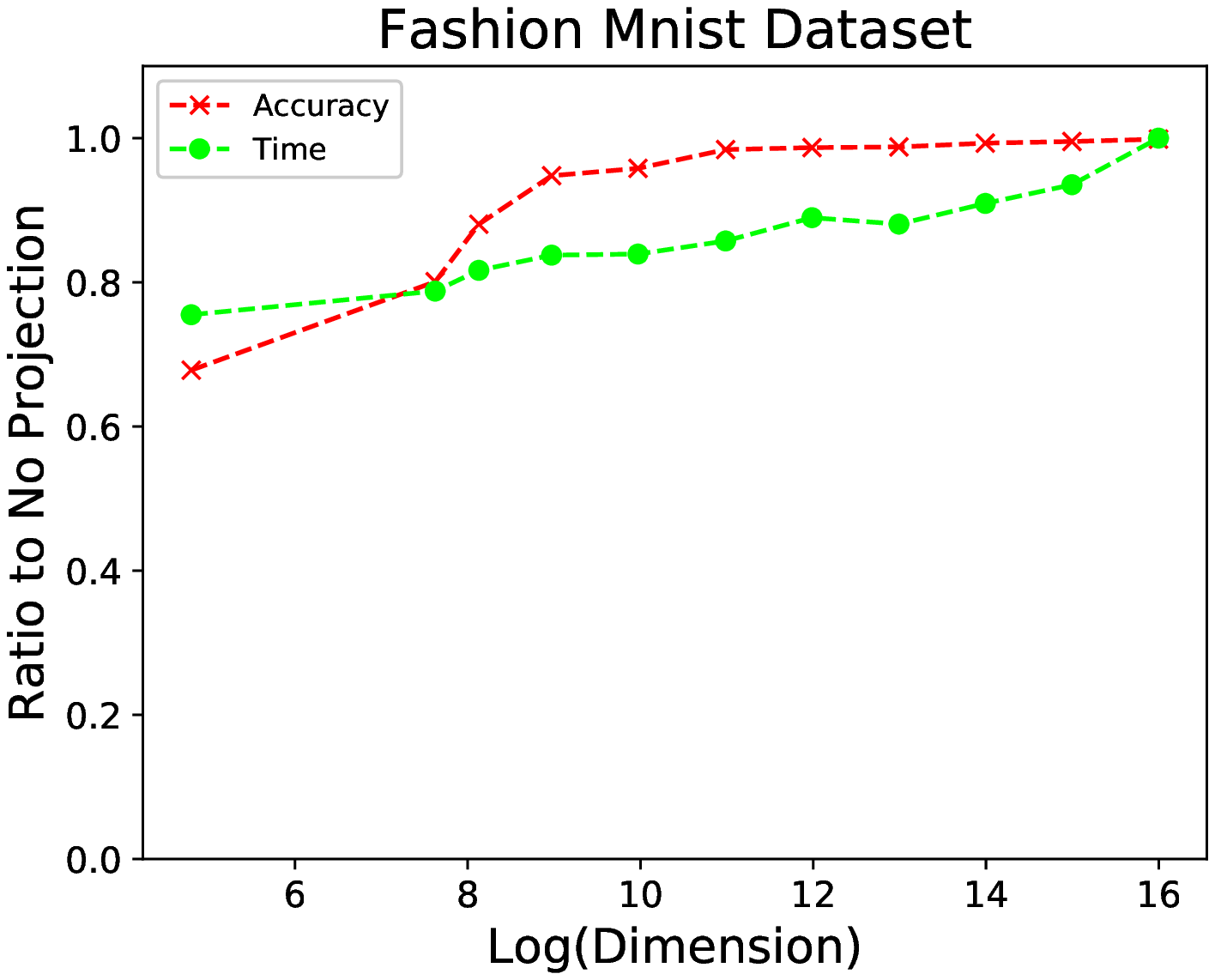}
\includegraphics[width=.4\textwidth]{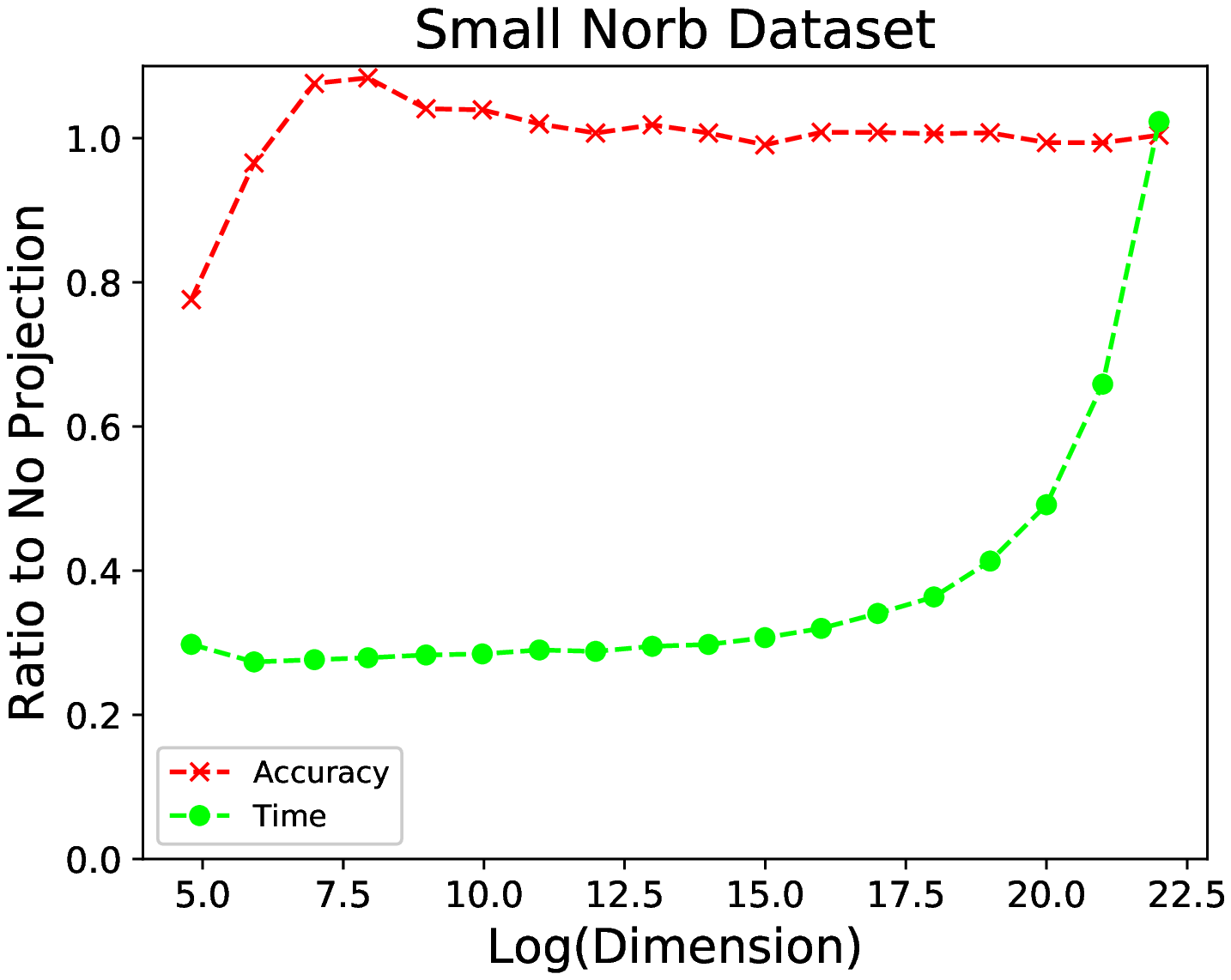}
\end{center}
  \caption{Green: Ratio of time taken to run t-SNE after dimensionality reduction to time taken to run t-SNE without dimensionality reduction. Red: Ratio of accuracy score after dimensionality reduction to accuracy score without dimensionality reduction. The t-SNE implementation used was openTSNE. $x$ axis shows the dimension after dimensionality reduction. The base of the logarithm is $1.5$. }
  \label{fig7} 
\end{figure}

We first compare the time taken to perform t-SNE and the accuracy scores achieved if a dimensionality reduction step is used before using t-SNE versus the case where no dimensionality reduction is used. We do so by plotting the ratio of the runtime and accuracy scores. For a given dimension, a higher accuracy score ratio is better and signifies that projecting down to that dimension using a random projection does not deteriorate the performance of t-SNE. Likewise, a lower time ratio is better. In Figure \ref{fig7} we plot the ratio of the accuracy scores and the time taken when we use the openTSNE implementation. The $x$ axis is the dimension of the random projection (ranging all the way up to the actual dimension of the input data). In all four datasets, we see that as the dimension increases, both the accuracy ratio and the ratio of the time taken approach $1$. However, we also observe that the accuracy score ratios approach $1$ much faster, indicating a ‘’sweet spot” where the dimension is high enough for the random projection to preserve geometric structure, while still low enough for t-SNE to run relatively quickly.

Likewise, we show the results of the same experiments using the scikit-learn implementation. Since the scikit-learn implementation is significantly slower than the openTSNE implementation, we subsample the $d$ values used and do not test the Small Norb dataset. However, even in this different implementation, we again observe the same trends as in the openTSNE implementation. 

\begin{figure}[htp]

\centering
\includegraphics[width=.33\textwidth]{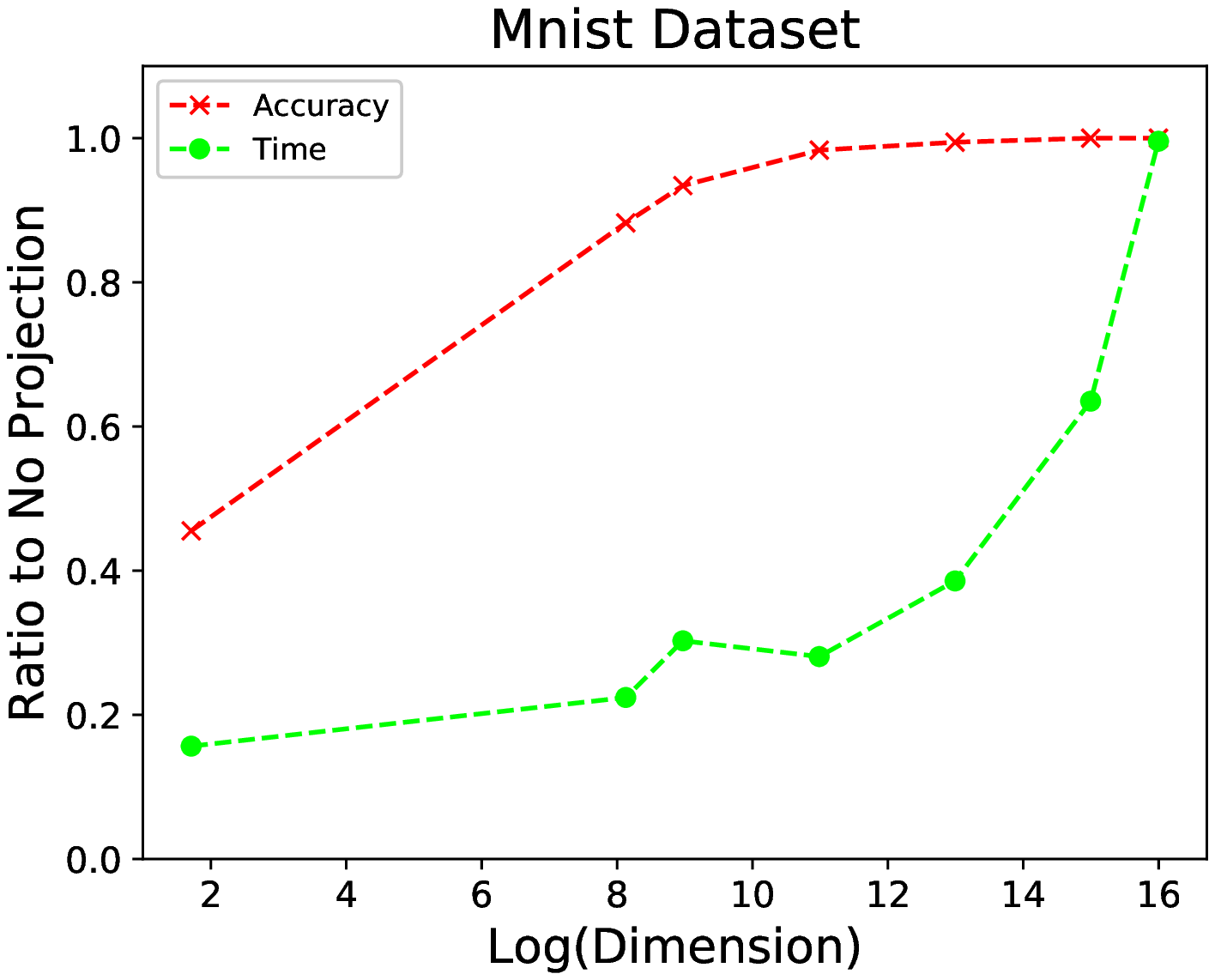}\hfill
\includegraphics[width=.33\textwidth]{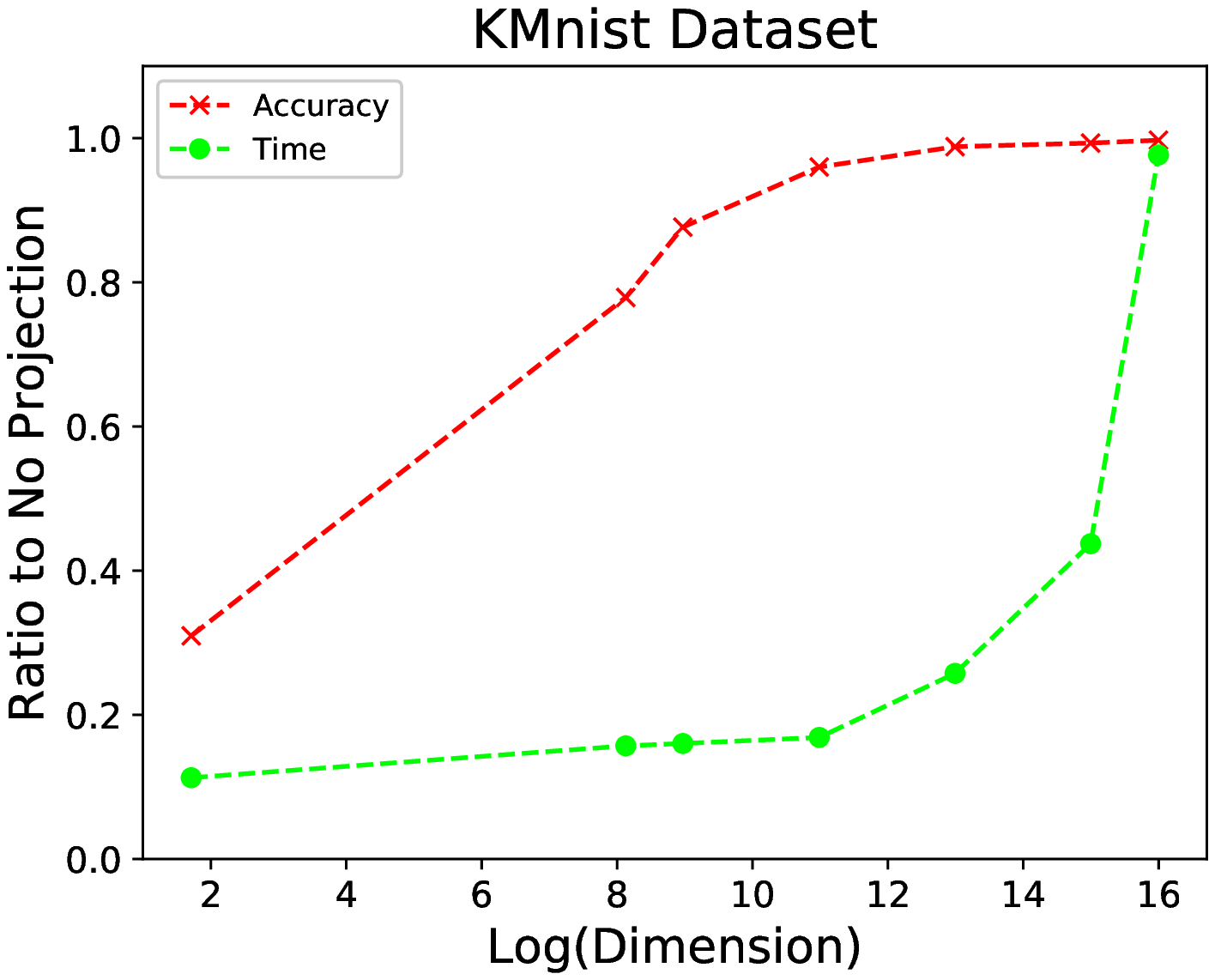}\hfill
\includegraphics[width=.33\textwidth]{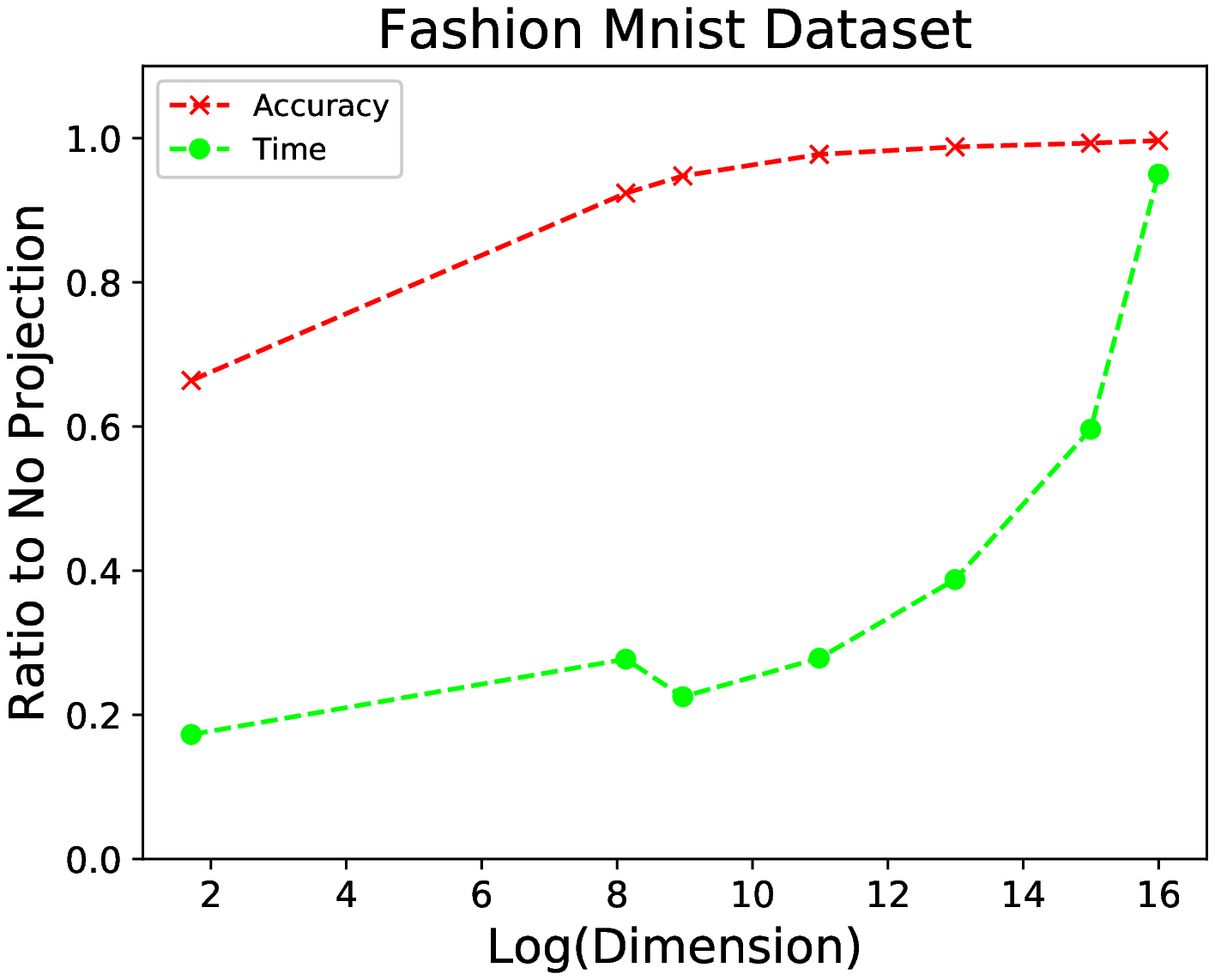}

\caption{Green: Ratio of time taken to run t-SNE after dimensionality reduction to time taken to run t-SNE without dimensionality reduction. Red: Ratio of accuracy score after dimensionality reduction to accuracy score without dimensionality reduction. The t-SNE implementation used was the scipy implementation. $x$ axis shows the dimension after dimensionality reduction. The base of the logarithm is $1.5$. Since the scipy implementation is very slow, we used less datapoints for the datasets shown and did not test on the Small Norb dataset.}
\label{fig:scipy}
\end{figure}

\begin{figure}[!htbp]
\floatbox[{\capbeside\thisfloatsetup{capbesideposition={right,top},capbesidewidth=7cm}}]{figure}[\FBwidth]
{\caption{Ratio of the accuracy score using PCA for dimensionality reduction instead of random projections and then performing t-SNE. We see that across all datasets, the accuracy scores become comparable to the scores when using t-SNE with no prior dimensionality reduction even for very small dimensions. This suggests that in practice, PCA might be a better choice than random projections to perform dimensionality reduction before t-SNE.}\label{fig:pca}}
{\includegraphics[width=6cm]{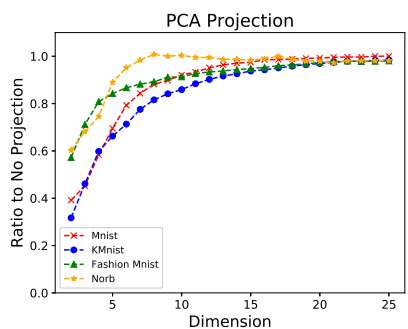}}
\end{figure}

In practice, SVD methods such as principal component analysis (PCA) are more widely used than random projections for the general task of dimensionality reduction. We show that for our four datasets, PCA indeed outperforms random projections. That is, we empirically observe that using PCA, rather than a random projection before t-SNE, allows us to project to a much smaller dimension while still retaining a high accuracy score ratio, (compared to the case where no dimensionality reduction is used). This is shown in Figure \ref{fig:pca} where regardless of the dataset, a projection to a dimension of $d = 25$ is sufficient to get an accuracy score ratio close to $1$. An advantage of random projections however, is that it is \emph{data oblivious} (the dimensionality reduction does not depend on the data) and it has provable guarantees in many cases, such as the JL lemma.

\begin{figure}[!htbp]
\floatbox[{\capbeside\thisfloatsetup{capbesideposition={right,top},capbesidewidth=4cm}}]{figure}[\FBwidth]
{\caption{Plotting the output of t-SNE.  Left: a dimensionality reduction to $d=50$ is performed before t-SNE on the MNIST dataset. Right: no dimensionality reduction is performed. Colors represent the label of the points. Observe that the same cluster structures appear in both plots.}\label{fig:clusterviz}}
{\includegraphics[width=.4\textwidth]{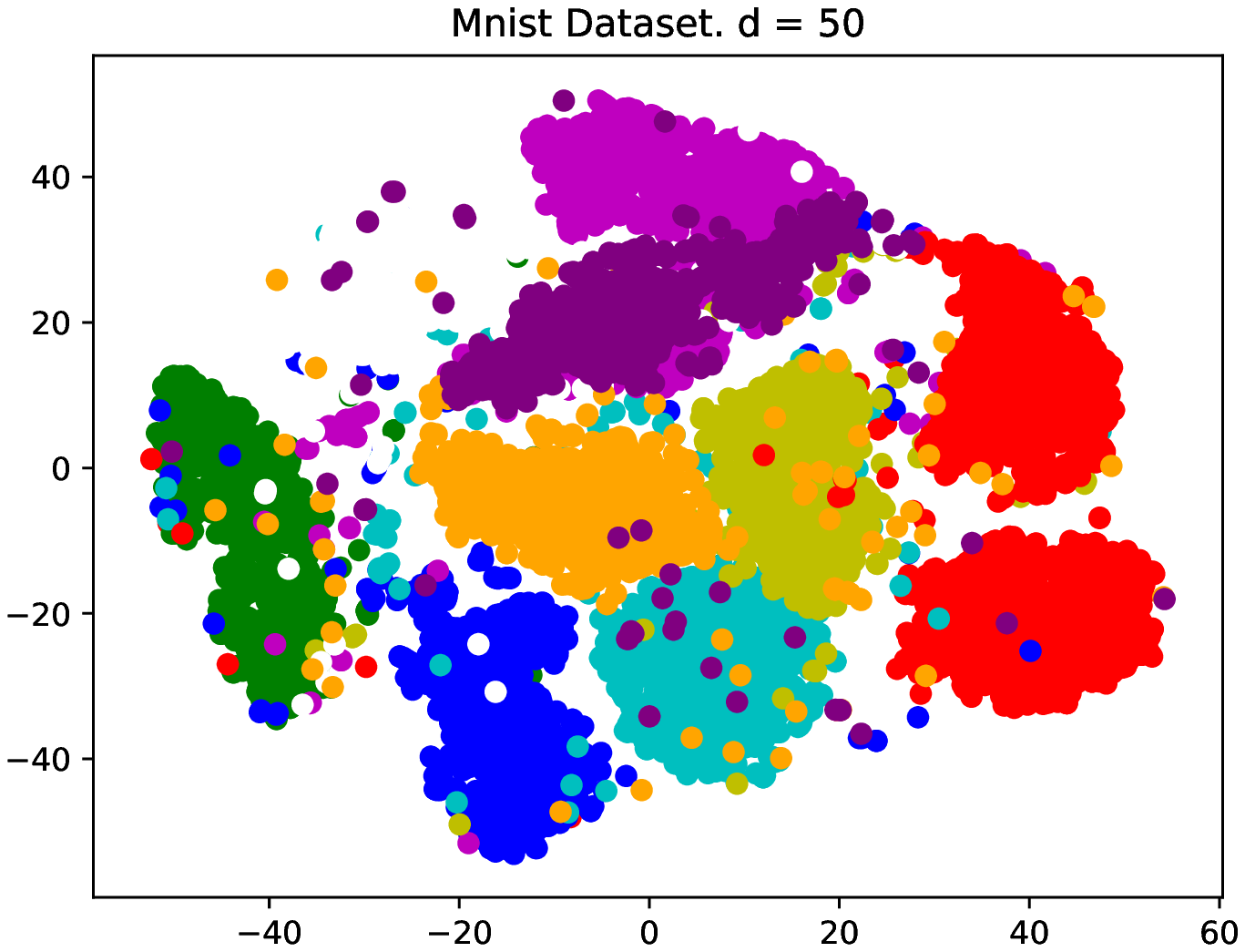}
\includegraphics[width=.4\textwidth]{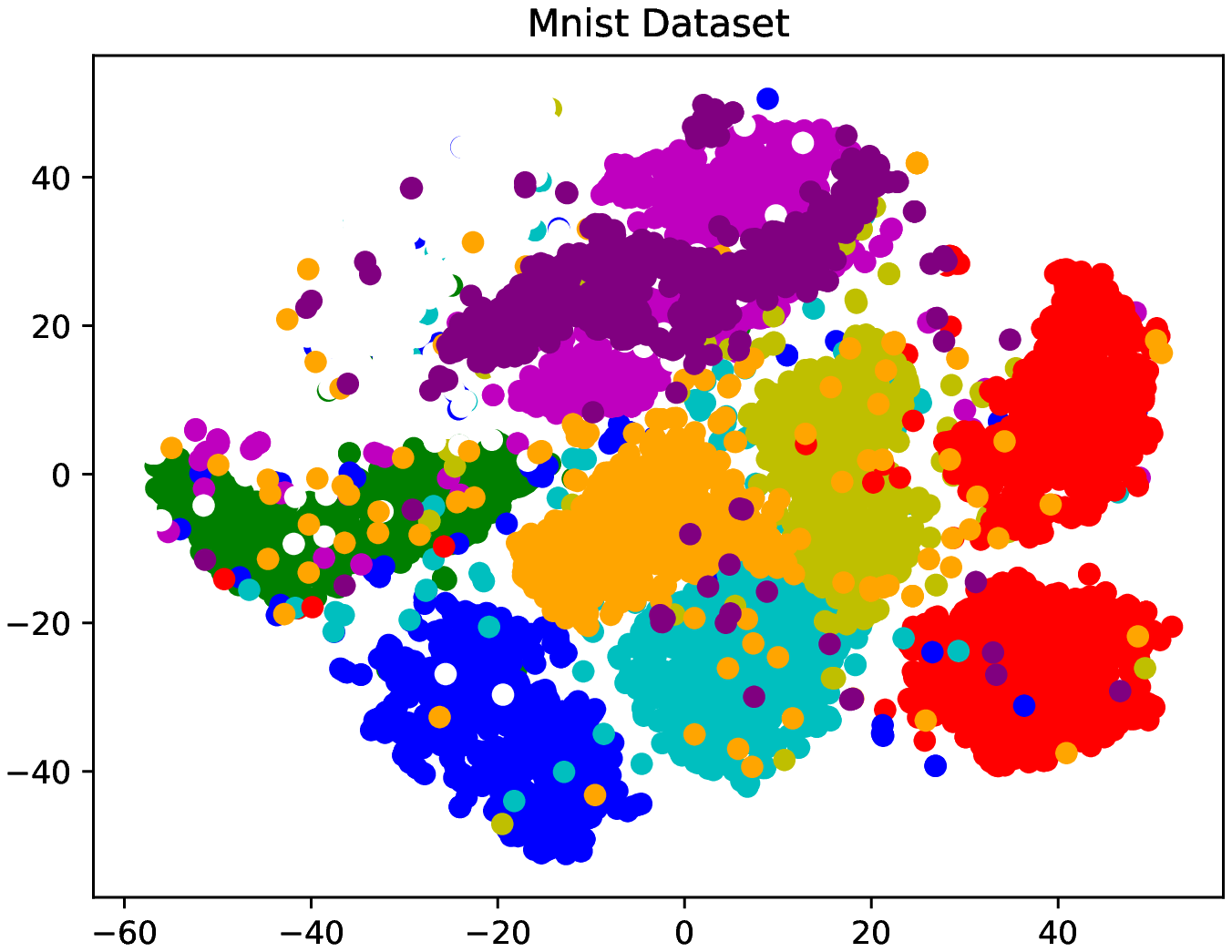}}
\end{figure}

\bibliography{paper}


\end{document}